\documentclass[letterpaper, 10 pt, conference]{ieeeconf}  

\IEEEoverridecommandlockouts                              
\usepackage{makecell}
\usepackage{graphicx}
\usepackage{subfig} 
\usepackage{url}

\usepackage{url}
\usepackage{breakurl}
\usepackage[breaklinks]{hyperref}
\usepackage{multirow}

\title{\LARGE \bf
Traffic Congestion Prediction Using Machine Learning Techniques
}

\author{Rafed Muhammad Yasir, Moumita Asad, Dr. Naushin Nower, Dr. Mohammad Shoyaib}

\begin{document}

\maketitle
\thispagestyle{empty}
\pagestyle{empty}

\begin{abstract}

The prediction of traffic congestion can serve a crucial role in making future decisions. Although many studies have been conducted regarding congestion, most of these could not cover all the important factors (e.g., weather conditions). We proposed a prediction model for the traffic congestion that can predict congestion based on day, time and several weather data (e.g., temperature, humidity). To evaluate our model, it has been tested against the traffic data of New Delhi. With this model, congestion of a road can be predicted one week ahead with an average RMSE of 1.12. Therefore, this model can be used to take preventive measure beforehand.

\end{abstract}

\section{INTRODUCTION}
Traffic congestion refers to a condition when travel demand exceeds the existing road system capacity \cite{rosenbloom1978peak}. It has become a major urban transportation problem \cite{downs2000stuck}, \cite{litman2009transportation}. It results in waste of valuable time, more fuel consumption, emission of pollutant gas, health hazards (lung diseases, high blood pressure) and low productivity at the workplace. World Bank's study revealed that around 8 billion USD are wasted every year in the  Greater Cairo Metropolitan Area(GCMA) due to traffic congestion \cite{WorldBank}. Drivers in Los Angeles spent more than 100 hours in traffic \cite{HiddenCostCongestion}. City dwellers in rich countries (e.g., New York, Los Angeles) lose approximately \$1,000 a year while sitting in traffic \cite{HiddenCostCongestion}. Therefore, a traffic prediction model is needed to take preventive measures to avoid it \cite{akbar2017predictive}.

Predicting traffic congestion is a challenging task since it shows nonlinear, time-varying characteristics \cite{deshpande2017performance}. In addition, many uncertain factors such as weather, time and day have an impact on traffic congestion, which makes it difficult to predict. Therefore, a model is required that incorporates all of these factors for predicting traffic congestion. Although a number of traffic prediction models have been proposed, most of the models could not cover all these factors. Furthermore, some of the models are only suitable for short-term prediction (e.g., 15 minutes). In this paper, we proposed a model by combining time, day and weather factors (e.g., temperature, rainfall) for long-term (one week ahead) traffic prediction. We validated the model using traffic data of New Delhi, India. The proposed model yielded an average RMSE of 1.12.

The rest of the paper is organized into different sections as follows. Section II presents on literature review. Section III discusses the methodology. Evaluation of the approach is discussed in section IV. Section V presents the concluding remarks.

\section{LITERATURE REVIEW}
Due to the increasing necessity for traffic predictive tools, various approaches have been proposed to predict traffic congestion \cite{min2007road}. Yisheng et al. proposed a deep-learning-based traffic flow prediction method by incorporating spatio-temporal relations (dependencies among traffic flows in space and time) \cite{lv2015traffic}. They used a stacked autoencoder (SAE) model to learn traffic flow patterns. The model was compared with four other models- back propagation neural network, random walk forecast method, support vector machine, and RBF neural network. To evaluate the effectiveness of each model, three performance indexes were used- MAE, RMSE, and MRE. Their proposed model achieved an RMSE of 50 for a 15-minute traffic flow prediction. However, the model yielded higher errors (e.g., RMSE=138.1) for longer time intervals (e.g., 45 mins).

Lee et al. incorporated weather data (e.g., rainfall, humidity, temperature) to predict traffic congestion \cite{lee2015prediction}. At first, they developed a multiple linear regression (MLR) model using 54 variables. Next, they filtered-out unimportant variables. The final model consists of 10 variables among which, 6 variables represent the days of the week and 4 are weather factors. Their approach yielded an accuracy of 75.5\%. However, they did not consider the time of the day, which is an important factor for predicting traffic congestion \cite{deshpande2017performance}.

Akbar et al. combined Complex Event Processing (CEP) with Machine Learning (ML) to predict traffic congestion \cite{akbar2017predictive}. They proposed an algorithm named Adaptive Moving Window Regression (AMWR) which uses SVR with radial based kernel function to predict intensity (number of vehicles per hour) and speed based on real-time and historical data. It uses Lomb Scargle method to find the optimum training window size. For ensuring a certain level of accuracy (80\%-95\%), AMWR adjusts the prediction window size. If the accuracy of the model is high (95\%\textgreater), the prediction window size is increased. On the other hand, the prediction window size is decreased when the accuracy of the model is low (\textless80\%). Lastly, the proposed technique uses CEP rules to predict traffic congestion based on the predicted intensity and speed. Although this technique achieved 96\% accuracy, the model mainly focuses on short-term traffic prediction.


\section{Methodology}
 In this paper, an approach was developed to predict traffic congestion one week ahead, based on the traffic and weather data of the previous week. For this purpose, the traffic and weather data of New Delhi, India, collected from the HERE API were used \cite{HereApi}. To train the model, Support Vector Regressor (SVR) algorithm was used since SVR achieved superior performance in traffic prediction than other methods \cite{deshpande2017performance}, \cite{ahn2016highway}.

\subsection{Data Collection} 
To predict traffic congestion, traffic data of New Delhi was collected using the HERE API \cite{HereApi}. The bounding box coordinates of the area, whose traffic data were collected, are (28.747193,77.091064) and (28.495247,77.304611). HERE also provides a weather api \cite{HereWeaterApi}, through which weather data of the selected region was collected. Following existing literature \cite{akbar2017predictive}, \cite{min2007road}, traffic and weather data was collected at an interval of 5 minutes. Table I summarizes the features present in the dataset.

\vspace{-2.2mm}
\begin{table}[hbt!]
\centering
\caption{Predictor and Response Classes, Names and Description}
\label{table:1}
\begin{tabular}{ |l|l|p{0.5\linewidth}| }
\hline
\textbf{Type} &	\textbf{Name} &	\textbf{Description} \\
\hline
\multirow{0}{*}{Predictor} & Source & Road from where the traffic is measured \\
& Destination & Road upto where traffic is  measured \\
& Time of a day & Time of the day when the data was collected \\
& Day of the week & Represents weekday \\
& Temperature & Temperature in degree celsius \\
& Daylight & Whether daylight existed or not \\
& Humidity & Humidity in percentage \\
& Wind speed & Wind Speed in Km/h \\
& Speed ratio & Ratio of current speed and average speed without traffic\\
\hline
\multirow{0}{*}{Response} &Jam factor & A number between 0.0 and 10.0 indicating the traffic level. As the number approaches  10.0 the quality of travel is getting worse. \\
\hline
\end{tabular}
\end{table}

\vspace{-2.1mm}
\subsection{Development of Prediction Model}
In this paper, an approach was developed to predict traffic congestion one week ahead, based on the traffic and weather data of the previous week. Several algorithms (e.g., LSTM, Random Forest Regressor, Gradient Boosting Regressor \cite{timeserml}) can be used for time series regression. We used support vector regression (SVR) since SVR achieved superior performance in traffic prediction than other methods \cite{deshpande2017performance}, \cite{ahn2016highway}. SVR is a supervised learning algorithm based on the computation of a linear regression function in a high dimensional feature \cite{basak2007support}. The input data is mapped to a higher dimension via a nonlinear function called kernel \cite{kernel}. Following an existing approach \cite{akbar2017predictive}, we used the radial basis function kernel (rbf) for data transformation.

\section{EXPERIMENTAL EVALUATION}
For evaluating our model, we collected traffic data of two consecutive weeks at an interval 5 minutes each day. The data of the first week (15.4.2019 to 21.4.2019) was used as training data and the second week (22.4.2019 to 28.4.2019) was used as testing data. To evaluate the performance of the proposed approach, four roads were selected randomly to measure Root Mean Square Error (RMSE), based on equation (1).

$$ RMSE = 
\Big[\frac{1}{n}\sum_{i=1}^n \Big(|f_i - \hat{f_i}|\Big)^2\Big]^\frac{1}{2} \eqno{(1)}
$$

\noindent where,  \\
\textit{f\textsubscript{i}} = the actual traffic congestion \\
\textit{\^{f\textsubscript{i}} }= the predicted traffic congestion

The predicted traffic congestion and the actual traffic congestion of the four roads are shown in Fig.1. Since we incorporated important factors (e.g.,weather, time and day) that have an impact on traffic congestion, it was expected that our model would yield low RMSE. However, the proposed approach could not achieve satisfactory result due to small dataset (1-week training data and 1-week testing data).

\begin{minipage}{\columnwidth}
    \makeatletter 
    \newcommand{\@captype}{figure}
    \makeatother
    \centering
    \subfloat[Location 1 (RMSE=0.893)]{%
      \includegraphics[width=0.95\textwidth]{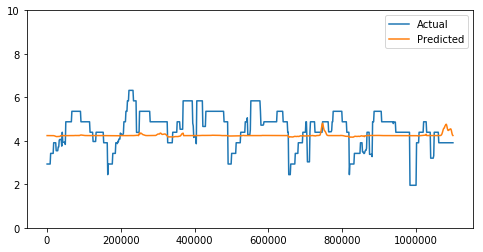}%
      \label{fig:location1}%
    }\qquad%
    \subfloat[Location 2 (RMSE=1.120)]{%
      \includegraphics[width=0.95\textwidth]{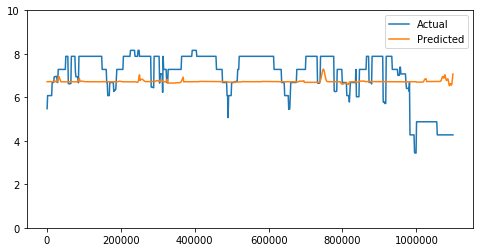}%
      \label{fig:location2}%
    }\qquad%
    \subfloat[Location 3 (RMSE=1.234)]{%
      \includegraphics[width=0.95\textwidth]{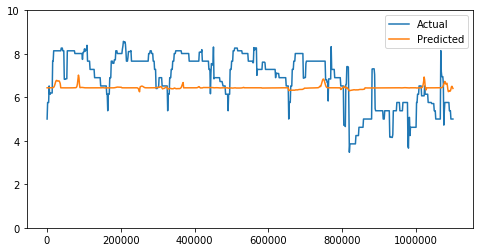}%
      \label{fig:location3}%
    }\qquad%
    \subfloat[Location 4 (RMSE=1.233)]{%
      \includegraphics[width=0.95\textwidth]{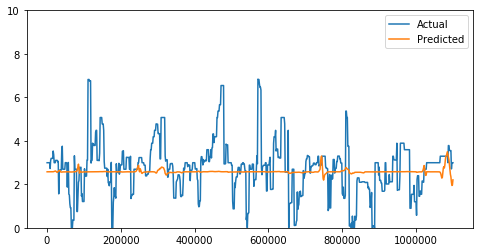}%
      \label{fig:location4}%
    }
    \caption{Prediction result on different roads}
\end{minipage}

\bigskip
To compare performance of the proposed approach with existing approaches, we implemented AMWR, which also uses SVR as its underlying method \cite{akbar2017predictive}. The results are shown in Fig.2 and Table II. Although AMWR achieved better result than our approach in most of the cases, it can  predict short-term traffic only (utmost next 15 minutes). On the other hand, the proposed technique can forecast traffic upto one week ahead.

\begin{figure}[h!]
  \includegraphics[width=0.49\textwidth]{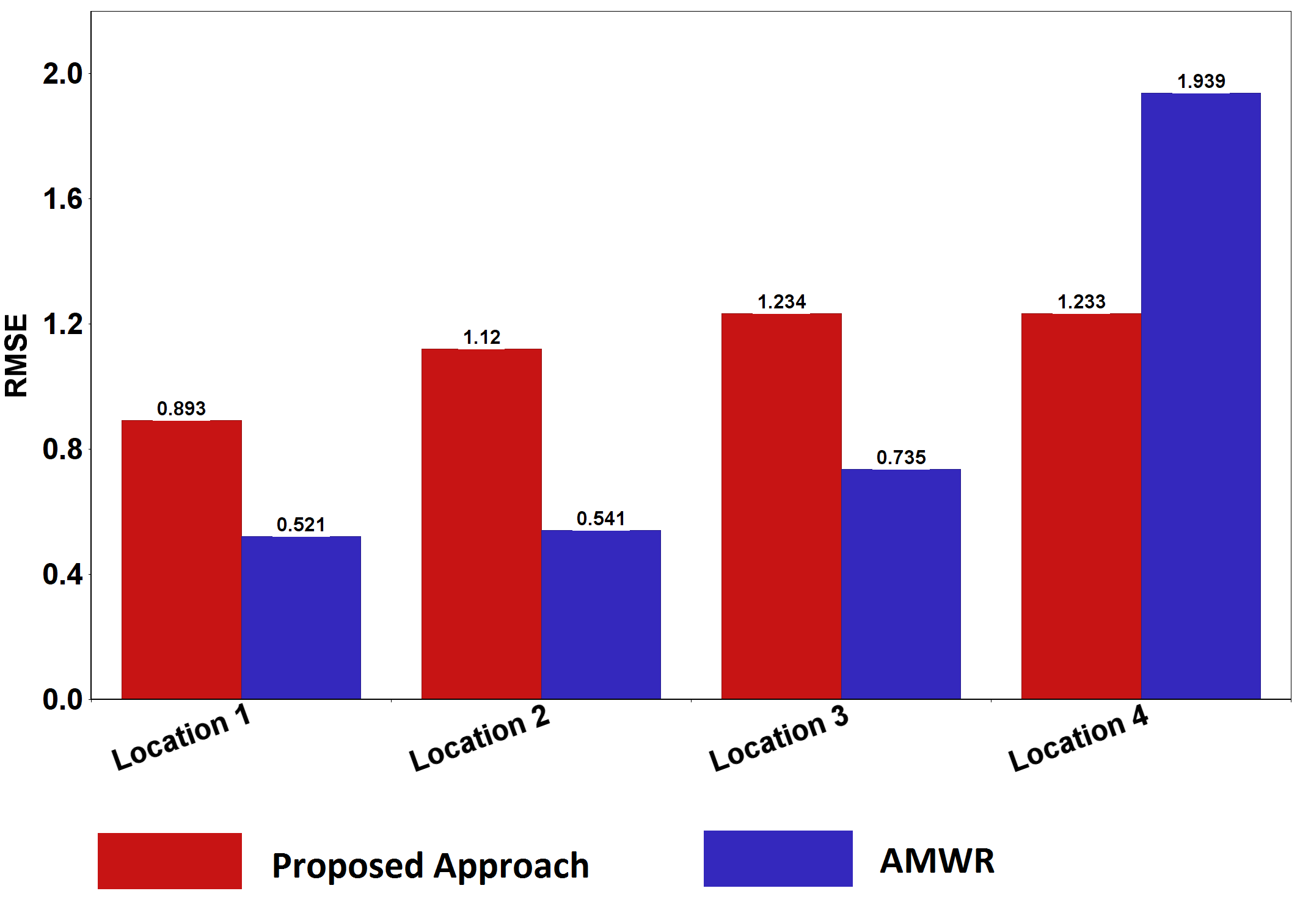}
  \caption{Comparison between Proposed Approach and AMWR}
\end{figure}

\begin{table}[h!]
\centering
\caption{Comparison between Proposed Approach and AMWR}
\label{table:2}
\begin{tabular}{|c|c|c|} 
\hline
\textbf{Location} & \textbf{RMSE of Proposed Approach} & \textbf{RMSE of AMWR} \\
\hline
Location 1 & 0.893 & 0.521 \\
\hline
Location 2 & 1.120 & 0.541  \\
\hline
Location 3 & 1.234 & 0.735 \\
\hline
Location 4 & 1.233 & 1.939 \\
\hline
\end{tabular}
\end{table}

\section{CONCLUSION}
In this paper, a traffic prediction model based on support vector regression with a radial basis kernel has been proposed. Apart from using features like road name, time and weekday, weather factors were also used for training the model as they are important factors that contribute to traffic congestion. The model was evaluated on four randomly selected road. It yielded an average RMSE of 1.12.

Due to small dataset, the model could not achieve expected accuracy. In future, accuracy of the model will be improved by increasing the size of the dataset. In addition, real time and historical data will be combined to further improve the accuracy.

\addtolength{\textheight}{-12cm}   





\bibliography{mybib}








\bibliographystyle{ieeetr}
\end{document}